\documentclass[runningheads,a4paper]{llncs}

\usepackage{amssymb}
\usepackage{graphicx}

\usepackage{url}
\usepackage{bm}
\usepackage{float}
\newcommand{\keywords}[1]{\par\addvspace\baselineskip
\noindent\keywordname\enspace\ignorespaces#1}

\urldef{\mailsa}\path|evgenyfo@post.tau.ac.il|
\urldef{\mailsb}\path|bron@eng.tau.ac.il|
\urldef{\mailsc}\path|michael.bronstein@usi.ch|

\begin{document}

\mainmatter  

\title{A correspondence-less approach to matching of deformable shapes}
\titlerunning{A correspondence-less approach to matching of deformable shapes}

%
%
\author{
Jonathan Pokrass$^1$ \and
Alexander M. Bronstein$^1$ \and
Michael M. Bronstein$^2$
}

\institute{
$^1$Dept. of Electrical Engineering, Tel Aviv University, Israel\\
\mailsa, \mailsb\\
$^2$Inst. of Computational Science, Faculty of Informatics,\\ Universit{\`a} della Svizzera Italiana, Lugano, Switzerland\\
\mailsc\\
}
\authorrunning{Pokrass {\em et al.}}


%
%

\toctitle{A correspondence-less approach to matching of deformable shapes}
\tocauthor{Pokrass {\em et al.}}
\maketitle

\begin{abstract}
Finding a match between partially available deformable shapes is a challenging problem with numerous applications. The problem is usually approached by computing local descriptors on a pair of shapes and then establishing a point-wise correspondence between the two.
In this paper, we introduce an alternative correspondence-less approach to matching fragments to an entire shape undergoing a non-rigid deformation. We use diffusion geometric descriptors and optimize over the integration domains on which the integral descriptors of the two
parts match. The problem is regularized using the Mumford-Shah functional. We show an efficient discretization based on the Ambrosio-Tortorelli approximation generalized to triangular meshes.
Experiments demonstrating the success of the proposed method are presented.

\keywords{deformable shapes, partial matching, partial correspondence, partial similarity, diffusion geometry, Laplace-Beltrami operator, shape descriptors, heat kernel signature, Mumford-Shah regularization}
\end{abstract}

\newcommand{\bb}[1]{\bm{\mathrm{#1}}}
\newcommand{\Tr}{\mathrm{T}}

\newcommand{\pp}{\bb{p}}
\newcommand{\qq}{\bb{q}}
\newcommand{\aaa}{\bb{a}}
\newcommand{\bbb}{\bb{b}}
\newcommand{\mm}{\bb{m}}
\newcommand{\nn}{\bb{n}}
\newcommand{\uu}{\bb{u}}
\newcommand{\vv}{\bb{v}}
\newcommand{\ww}{\bb{w}}
\newcommand{\xx}{\bb{x}}
\newcommand{\yy}{\bb{y}}
\newcommand{\ff}{\bb{f}}
\newcommand{\gggg}{\bb{g}}

\newcommand{\mmu}{\bb{\mu}}
\newcommand{\nnu}{\bb{\nu}}
\newcommand{\rrho}{\bb{\rho}}
\newcommand{\ssigma}{\bb{\sigma}}
\newcommand{\ttau}{\bb{\tau}}
\newcommand{\ttheta}{\bb{\theta}}
\newcommand{\bbeta}{\bb{\beta}}

\newcommand{\ones}{\bb{1}}
\newcommand{\zeros}{\bb{0}}

\newcommand{\Ee}{\bb{E}}
\newcommand{\Pp}{\bb{P}}
\newcommand{\Qq}{\bb{Q}}
\newcommand{\Aa}{\bb{A}}
\newcommand{\Bb}{\bb{B}}
\newcommand{\Mm}{\bb{M}}
\newcommand{\Nn}{\bb{N}}
\newcommand{\Xx}{\bb{X}}
\newcommand{\Gg}{\bb{G}}
\newcommand{\Ii}{\bb{I}}
\newcommand{\Dd}{\bb{D}}
\newcommand{\Ww}{\bb{W}}
\newcommand{\Rr}{\bb{R}}
\newcommand{\Ss}{\bb{S}}

\newcommand{\mypara}[1]{{\noindent \bf{#1.} }}

\section{Introduction}


In many real-world settings of the shape recognition problem, the data are degraded by acquisition imperfections and noise, resulting in the need to find {\em partial} similarity of objects.
Such cases are common, for example, in face recognition, where the facial surface may be partially occluded by hair. 
In other applications, such as shape retrieval, correct semantic similarity of two objects is based on partial similarity -- 
for example, a centaur is partially similar to a human because they share the human-like upper body \cite{jacobs:centaur}.


In rigid shape analysis, modifications of the popular iterative closest point (ICP) algorithm are able to deal with partial shape alignment by rejecting points with bad correspondences (e.g., by thresholding the product of local normal vectors).
However, it is impossible to guarantee how large and regular the resulting corresponding parts will be.

Bronstein {\em et al.} \cite{bronstein2009partial} formulated non-rigid partial similarity as a multi-criterion optimization problem, in which one tries to find the  corresponding parts in two shapes by simultaneously maximizing significance and similarity criteria (in  \cite{bronstein2009partial}, metric distortion \cite{gro:GEOMETRY,mem:sap1:GEOMETRY,bro:bro:kim:PNAS} was used as a criterion of similarity, and part area as significance).
The problem requires the knowledge of correspondence between the shapes, and in the absence of a given correspondence, can be solved by alternating between weighted correspondence finding and maximization of part area.
In \cite{bronstein2009partial}, a different significance criterion based on statistical occurrence of local shape descriptors was used.

One of the drawbacks of the above method is its tendency in some cases to find a large number of disconnected components, which have the same area as a larger single component.
The same authors addressed this problem using a Mumford-Shah \cite{mumford1989optimal,chan2001active}-like regularization  for rigid \cite{bronstein2008regularized} and non-rigid \cite{bronstein2008not} shapes.

Recent works on local shape descriptors (see, e.g.,
descriptors \cite{johnson1999usi,moments:zhang2001efe,pauly2003msf,clarenz2004rfd,manay2004iis,mitra2006pfs,zaharescu-surface,sun2009concise,SB10b,Iasonas}) have led to the adoption of bags of features \cite{siv:zis:CVPR:03} approach popular in image analysis for the description of 3D shapes \cite{mitra2006pfs,ovsjanikov2009shape,TolCasFus09}. Bags of features allows to some extent finding partial similarity, if the overlap between the parts is sufficiently large.

In this paper, we present an approach for correspondence-less partial matching of non-rigid 3D shapes.
Our work is inspired by the recent work on partial matching of images \cite{domokos2010affine}. The main idea of this approach, adopted here, is to find similar parts by comparing part-wise distributions of local descriptors. This removes the need of correspondence knowledge and greatly simplifies the problem.

The rest of the paper is organized is as follows.
In Section 2, we review the mathematical background of diffusion geometry, which is used for the construction of local descriptors.
Section 3 deals with the partial matching problem and Section 4 addresses its discretization.
Section 5 presents experimental results.
Finally, Section 6 concludes the paper.

\section{Background}
\label{sec:backgr}

{\bf Diffusion geometry.}
Diffusion geometry is an umbrella term referring to geometric analysis of diffusion or random walk processes.
We models a shape as a compact two-dimensional Riemannian manifold $X$.
In it simplest setting, a diffusion process on $X$ is described by the partial differential equation
\begin{eqnarray}
\left(\frac{\partial}{\partial t} + \Delta \right)f(t,x) = 0,
\label{eq:heat}
\end{eqnarray}
called the \emph{heat equation}, where $\Delta$ denotes the positive-semidefinite Laplace-Beltrami operator associated with the Riemannian metric of $X$.
The heat equation describes
the propagation of heat on the surface and its solution $f(t,x)$ is the heat distribution at a point $x$ in time $t$.
The initial condition of the equation is some initial heat distribution $f(0,x)$;
if $X$ has a boundary, appropriate boundary conditions must be added.

%
The solution of~(\ref{eq:heat}) corresponding to a point initial condition
 $f(0,x) = \delta(x,y)$, is called the {\em heat kernel} and represents the amount of heat
 transferred from $x$ to $y$ in time $t$ due to the diffusion process.
 The value of the heat kernel $h_t(x,y)$ can also be interpreted as the transition probability
density of a random walk of length $t$ from the point $x$ to the point $y$.

Using spectral decomposition, the heat kernel can be represented as
\begin{eqnarray}
h_t(x,y) &=& \sum_{i\geq 0} e^{-\lambda_i t} \phi_i(x) \phi_i(y).
\label{eq:heatkernel}
\end{eqnarray}
Here, $\phi_i$ and $\lambda_i$ denote, respectively, the eigenfunctions and eigenvalues of the Laplace-Beltrami operator
 satisfying $\Delta \phi_i = \lambda_i \phi_i$ (without loss of generality, we assume $\lambda_i$ to be sorted in increasing order starting with
 $\lambda_0 = 0$).
Since the Laplace-Beltrami operator is an {\em intrinsic} geometric quantity, i.e., it can be expressed solely in terms of the
 metric of $X$, its eigenfunctions and eigenvalues as well as the heat kernel are invariant under isometric transformations (bending) of
 the shape.


{\bf Heat kernel signatures. }
By setting $y=x$, the heat kernel $h_t(x,x)$ expresses the probability density of remaining at a point $x$  after time $t$.
The value $h_t(x,x)$, sometimes referred to as the \emph{auto-diffusivity function}, is related to the Gaussian curvature $K(x)$ through
\begin{eqnarray}
h_t(x,x) &\approx & \frac{1}{4\pi t}\left(1 + \frac{1}{6}K(x) t + \mathcal{O}(t^2)\right).
\end{eqnarray}
This relation coincides with the well-known fact
that heat tends to diffuse slower at points with positive
curvature, and faster at points with negative curvature.
Under mild technical conditions, the set $\{ h_t(x,x) \}_{t > 0}$ is fully informative in the sense that
it allows to reconstruct the Riemannian metric of the manifold \cite{sun2009concise}.

Sun \emph{et al.} \cite{sun2009concise} proposed constructing point-wise descriptors referred to as {\em heat kernel signatures} (HKS) by taking the values of the discrete auto-diffusivity function at point $x$ at multiple times, $\pp(x) = c(x)(h_{t_1}(x,x),..., h_{t_d}(x,x))$, where $t_1,..., t_d$ are some fixed time values and $c(x)$ is chosen so that $||p(x)||_2=1$ .
Such a descriptor is a vector of dimensionality $d$ at each point.
Since the heat kernel is an intrinsic quantity, the HKS is invariant to isometric transformations of the shape.
%

A scale-invariant version of the HKS descriptor (SI-HKS) was proposed in \cite{Iasonas}.
First, the heat kernel is sampled logarithmically in time.
Next, the logarithm and a derivative with respect to time of the heat kernel values are taken to undo the multiplicative constant.
Finally, taking the magnitude of the Fourier transform allows to undo the scaling of the time variable.

%
{\bf Bags of features.} Ovsjanikov {\em et al.} \cite{ovsjanikov2009shape} and Toldo {\em et al.} \cite{TolCasFus09} proposed constructing global shape descriptors from local descriptors
using the {\em bag of features} paradigm \cite{siv:zis:CVPR:03}.
In this approach, a fixed ``geometric vocabulary''
is computed by means of an off-line clustering of the descriptor space.
Next, each point descriptor 
is represented in the vocabulary using vector quantization. 
%
The bag of features global shape descriptor is then computed as the histogram of quantized descriptors over the entire shape.


\section{Partial matching}

In what follows, we assume to be given two shapes $X$ and $Y$ with corresponding point-wise descriptor fields $\pp$ and $\qq$ defined on them (here we adopt HKS descriptors, though their quantized variants or any other intrinsic point-wise descriptors can be used as well).
Assuming that $Y$ is a part of an unknown shape that is intrinsically similar to $X$, we aim at finding a part $X' \subseteq X$
having the same area $A$ of $Y$ such that the integral shape descriptors computed on $X'$ and $Y$ coincide as closely as possible.
In order to prevent the parts from being fragmented and irregular, we penalize for their boundary length. The entire problem can be expressed as minimization of the following energy functional
\begin{eqnarray}
	E(X') &=& \left\| \int_{X'} \pp da - \overline{\qq} \right\|^2 + \lambda_\mathrm{r} L(\partial X')
\label{eq:orig-min-prob}
\end{eqnarray}
under the constraint $A(X') = A$, where $A$ denotes area and $\displaystyle{\overline{\qq} = \int_{Y} \qq da}$. The first term of the functional constitutes
the data term while the second one is the regularity term whose influence is controlled by the parameter $\lambda_\mathrm{r}$.

Discretization of the above minimization problem with a crisp set $X'$ results in combinatorial complexity. To circumvent this difficulty, in \cite{bronstein2008not,bronstein2008regularized} it was proposed to relax the problem by replacing the crisp part $X'$ by a fuzzy membership function $u$ on $X$, replacing the functional $E$ by a generalization of the Mumford-Shah functional \cite{mumford1989optimal} to surfaces.
Here, we adopt this relaxation as well as the approximation of the Mumford-Shah functional proposed by Ambrosio and Tortorelli \cite{ambrosio1990approximation}. This yields the problem of the form
\begin{eqnarray}
\lefteqn{\min_{u,\rho,\sigma} D(u) +  \lambda_\mathrm{r}R(u;\rho) }\nonumber\\
&& \mathrm{s.t.} \int_X u da = A,
\label{eq:min-ambrosio-tortorelli}
\end{eqnarray}
with the data term
\begin{eqnarray}
D (u) &=& \left\| \int_X \pp u da - \overline{\qq} \right\|^2
\label{eq:cont-data}
\end{eqnarray}
and the Ambrosio-Tortorelli regularity term
\begin{eqnarray}
R(u;\rho) &=& \frac{\lambda_\mathrm{s}}{2} \int_X \rho^2 \| \nabla u \|^2 da
+ \lambda_\mathrm{b} \epsilon \int_X \| \nabla \rho \|^2 da
+ \frac{\lambda_\mathrm{b}}{4\epsilon} \int_X (1-\rho)^2 da,
\label{eq:cont-reg}
\end{eqnarray}
where $\rho$ is the so-called phase field indicating the discontinuities of $u$, and $\epsilon > 0$ is a parameter.

The first term of $R$ above imposes piece-wise smoothness of the fuzzy part $u$  governed by the parameter $\lambda_\mathrm{s}$. By setting a sufficiently large $\lambda_\mathrm{s}$, the parts become approximately piece-wise constant as desired in the original crisp formulation (\ref{eq:orig-min-prob}).
The second term of $R$ is analogous to the boundary length term in (\ref{eq:orig-min-prob}) and converges to the latter as $\epsilon \rightarrow 0$.

We minimize (\ref{eq:min-ambrosio-tortorelli}) using alternating minimization comprising the following two iteratively repeated steps:

\noindent {\bf Step 1:} fix $\rho$ and solve for $u$
\begin{eqnarray}
\min_{u} \left\| \int_X \pp u da - \overline{\qq} \right\|^2 + \lambda_\mathrm{r}\frac{\lambda_\mathrm{s}}{2} \int_X \rho^2 \| \nabla u \|^2 da &\,\,\mathrm{s.t.}\,\,& \int_X u da = A.
\label{eq:altmin-cont-step1}
\end{eqnarray}

\noindent {\bf Step 2:} fix the part $u$ and solve for $\rho$
\begin{eqnarray}
\min_{\rho} \frac{\lambda_\mathrm{s}}{2} \int_X \rho^2 \| \nabla u \|^2 da
+ \lambda_\mathrm{b} \epsilon \int_X \| \nabla \rho \|^2 da
+ \frac{\lambda_\mathrm{b}}{4\epsilon} \int_X (1-\rho)^2 da.
\label{eq:altmin-cont-step2}
\end{eqnarray}


\section{Discretization and numerical aspects}

We represent the surface $X$ as triangular mesh with $n$ faces constructed upon the samples $\{\xx_1,\dots,\xx_m\}$
and denote by $\aaa = (a_1,\dots,a_m)^\Tr$ the corresponding area elements at each vertex. 
$\Aa = \mathrm{diag}\{\aaa\}$ denote the diagonal $m \times m$ matrix created out of $\aaa$.
The membership function $u$ is sampled at each vertex and represented as the vector $\uu = (u_1,\dots,u_m)^\Tr$.
Similarly, the phase field is represented as the vector $\rrho = (\rho_1,\dots,\rho_m)^\Tr$.

{\bf Descriptors.} 
The computation of the discrete heat kernel $h_t(\xx_1, \xx_2)$ requires computing discrete eigenvalues and eigenfunctions of the discrete Laplace-Beltrami operator.
The latter can be computed directly using the finite elements method (FEM) \cite{reuter},
of by discretization of the Laplace operator on the mesh followed by its eigendecomposition.
Here, we adopt the second approach according to which the discrete Laplace-Beltrami operator is expressed in the following generic form,
\begin{equation}
\label{eq:ldb_discrete}
(\Delta f)_i = \frac{1}{a_i} \sum_{j} w_{ij} (f_i - f_j),
\end{equation}
where $f_i = f(\xx_i)$ is a scalar function defined on the mesh, $w_{ij}$ are weights, and $a_i$ are
normalization coefficients.
In matrix notation, (\ref{eq:ldb_discrete}) can be written as
$\Delta \ff = \Aa^{-1} \Ww \ff$,
where $\ff$ is an $m \times 1$ vector and $\Ww = \mathrm{diag}\left\{\sum_{l\neq i} w_{il} \right\} - w_{ij}$.

The discrete eigenfunctions and eigenvalues are found by solving the \emph{generalized eigendecomposition} \cite{levy2006lbe}
$\Ww \bm{\mathrm{\Phi}} = \Aa \bm{\mathrm{\Phi}} \bm{\mathrm{\Lambda}}$,
where $\bm{\mathrm{\Lambda}} = \mathrm{diag}\{\lambda_l\}$ is a diagonal matrix of eigenvalues and $\bm{\mathrm{\Phi}} = (\phi_l(x_i))$
is the matrix of the corresponding eigenvectors.

Different choices of $\Ww$ have been studied,
depending on which continuous properties of the Laplace-Beltrami operator one
wishes to preserve \cite{floater2005surface,wardetzky2008discrete}. 
For triangular meshes, a popular choice adopted in this paper is
the \emph{cotangent weight} scheme \cite{pinkall1993computing,meyer2003ddg},
in which
\begin{equation}
w_{ij} = \left\{
\begin{array}{ll}
(\cot \beta_{ij} + \cot \gamma_{ij})/2 & :~ \xx_j \in \mathcal{N}(\xx_j);\\
0 & :~ else,
\end{array}
\right.
\end{equation}
where $\beta_{ij}$ and $\gamma_{ij}$ are the two
angles opposite to the edge between vertices $\xx_i$ and $\xx_j$ in the two triangles sharing the edge.

{\bf Data term. }
Denoting by $\Pp$ $d \times m$ the matrix of point-wise descriptors on $X$ (stored in columns), we have
\begin{eqnarray}
\int \pp u da & \approx & \Pp \mathrm{diag}\{\Aa\} \uu
\end{eqnarray}
This yields the following discretization of the data term (\ref{eq:cont-data})
\begin{eqnarray}
D(\uu) &=& \| \Pp \Aa \uu - \overline{\qq} \|^2 = \uu^\Tr \Aa^\Tr \Pp^\Tr \Pp \Aa \uu - 2 \overline{\qq}^\Tr \Pp \Aa \uu + \overline{\qq}^\Tr \overline{\qq}.
\end{eqnarray}

{\bf Gradient norm. }
We start by deriving the discretization of a single term $\rho^2 \| \nabla u \|^2 da$ in some triangle $t$ of the mesh.
Let us denote by $\xx_i,\xx_j$ and $\xx_k$ the vertices of the triangle and
let $\Xx_t = (\xx_j - \xx_i, \xx_k-\xx_i)$ be the $3\times 2$ matrix whose columns are the vectors forming the triangle, and by
$\alpha_t = \frac{1}{2}\sqrt{\det (\Xx_t^\Tr \Xx_t)}$ its area.
Let also $\Dd_t$ be the sparse $2 \times m$ matrix with $+1$ at indices $(1,j)$ and $(2,k)$, and $-1$ at $(1,i)$ and $(2,i)$.
$\Dd_t$ is constructed in such a way to give the differences of values of $u$ on the vertices of the triangle with respect to the
values at the central vertex, $\Dd_t \uu = (u_j - u_i, u_k-u_i)^\Tr$.
The gradient of the function $u$ is constant on the triangle and can be expressed in these terms by $\gggg_t = (\Xx_t^\Tr \Xx_t)^{-1/2} \Dd_t \uu = \Gg_t \uu$.

In order to introduce the weighting by $\rho^2$, let $\Ss$ be an $n \times m$ sparse matrix with the elements $s_{ti}=\frac{1}{3}$ for every vertex $i$ belonging to the triangle $t$ and zero otherwise.
In this notation, $\Ss\rrho$ is a per-triangle field whose elements are the average values of $\rrho$ on each of the mesh triangles.
We use the Kroenecker product of $\Ss$ with $\ones = (1,1)^\Tr$ to define the $2n \times n$ matrix $\Ss \otimes \ones$ formed by replicating twice each of the rows of $\Ss$.
This yields
\begin{eqnarray}
	\lefteqn{\int \rho^2 \| \nabla u\|^2 da \approx  \sum_{t } \| (\Ss \rrho)_t \Gg_t \uu \|^2\alpha_t = } \nonumber\\
&& = \left\| \mathrm{diag}\{ (\Ss \otimes \ones) \rrho \} \left( \begin{array}{c}\sqrt{\alpha_1}\Gg_1  \\
             \vdots \\
           \sqrt{\alpha_n}\Gg_n
           \end{array}
 \right)\right\|^2 = \uu^\Tr \Gg^\Tr \Ss(\rrho) \Gg \uu,
 \label{eq:gradu_rho}
\end{eqnarray}
where $\Gg$ is the matrix containing $\sqrt{\alpha_t}\Gg_t$ stacked as rows, and $\Ss(\rrho) = \mathrm{diag}\{ (\Ss \otimes \ones) \rrho \}^2$.

{\bf Discretized alternating minimization. }
We plug in the results obtained so far into the two steps of the alternating minimization problem (\ref{eq:altmin-cont-step1})--(\ref{eq:altmin-cont-step2}).
For fixed $\rrho$,
the discretized minimization problem (\ref{eq:altmin-cont-step1}) w.r.t. $\uu$  can be written as
\begin{eqnarray}
\min_{\uu} \uu^\Tr \left(\Aa^\Tr \Pp^\Tr \Pp \Aa + \lambda_\mathrm{r}\frac{\lambda_\mathrm{s}}{2}  \Gg^\Tr \Ss(\rrho) \Gg \right)  \uu - 2 \overline{\qq}^\Tr \Pp \Aa \uu
& ~~~\mathrm{s.t.}~~~ & \aaa^\Tr \uu = A
\end{eqnarray}

Let us now fix $\uu$. In a triangle $t$, we denote $g_t^2 = \| \Gg_t \uu \|^2$ and let $\Rr(\uu) = \mathrm{diag}\{\alpha_1 g_1^2, \dots, \alpha_n g_n^2 \}$.
Using this notation, we obtain the following discretization of the integrals in the regularization term (\ref{eq:cont-reg})
\begin{eqnarray}
\int_X \rho^2 \| \nabla u\|^2 da & \approx & \sum_{t} (\Ss \rho)^2_t g_t^2\alpha_t = \rrho^\Tr \Ss^\Tr \Rr(\uu) \Ss \rrho.
\end{eqnarray}

Similar to the derivation of (\ref{eq:gradu_rho}),
\begin{eqnarray}
\int \| \nabla \rho \|^2 da  & \approx & \rrho^\Tr \Ss \Gg^\Tr \Gg \Ss \rrho,
\end{eqnarray}
and
\begin{eqnarray}
\int_X (1-\rho)^2 da  &\approx &\rrho^\Tr \Aa \rrho - 2\aaa^\Tr \rrho  + \ones^\Tr \aaa.
\end{eqnarray}
The discretized minimization problem (\ref{eq:altmin-cont-step2}) w.r.t. $\rrho$ becomes
\begin{eqnarray}
\min_{\rrho} \rrho^\Tr\left(  2\epsilon \lambda_\mathrm{s} \Ss^\Tr \Rr(\uu) \Ss  + 4 \epsilon^2 \lambda_\mathrm{b} \Ss \Gg^\Tr \Gg \Ss + \lambda_\mathrm{b} \Aa \right)\rrho - 2\lambda_\mathrm{b} \aaa^\Tr \rho.
\end{eqnarray}
Since the above is an unconstrained quadratic problem, it has the following closed-form solution
\begin{eqnarray}
\rrho &=&\left(  \frac{2\epsilon \lambda_\mathrm{s}}{\lambda_\mathrm{b}} \Ss^\Tr \Rr(\uu) \Ss  + 4 \epsilon^2 \Ss \Gg^\Tr \Gg \Ss + \Aa \right)^{-1} \aaa.
\end{eqnarray}

\section{Results}

In order to test our approach, we performed several partial matching experiments
on data from the SHREC 2010 benchmark \cite{bronstein2010shrec1,bronstein2010shrec2}
and the TOSCA dataset \cite{bronstein2008ngn}.\footnote{Both datasets are available online at \url{http://tosca.cs.technion.ac.il}} The datasets contained high-resolution (10K-50K vertices) triangular meshes of human and animal shapes with simulated transformations, with known groundtruth correspondence between the transformed shapes.
In our experiments, all the shapes were downsampled to approximately 2500 vertices.
Parts were cut by taking a geodesic circle of random radius around a random center point.

\begin{figure}[tbh]
\begin{small}
    \begin{center}
    \includegraphics[width=0.6\linewidth]{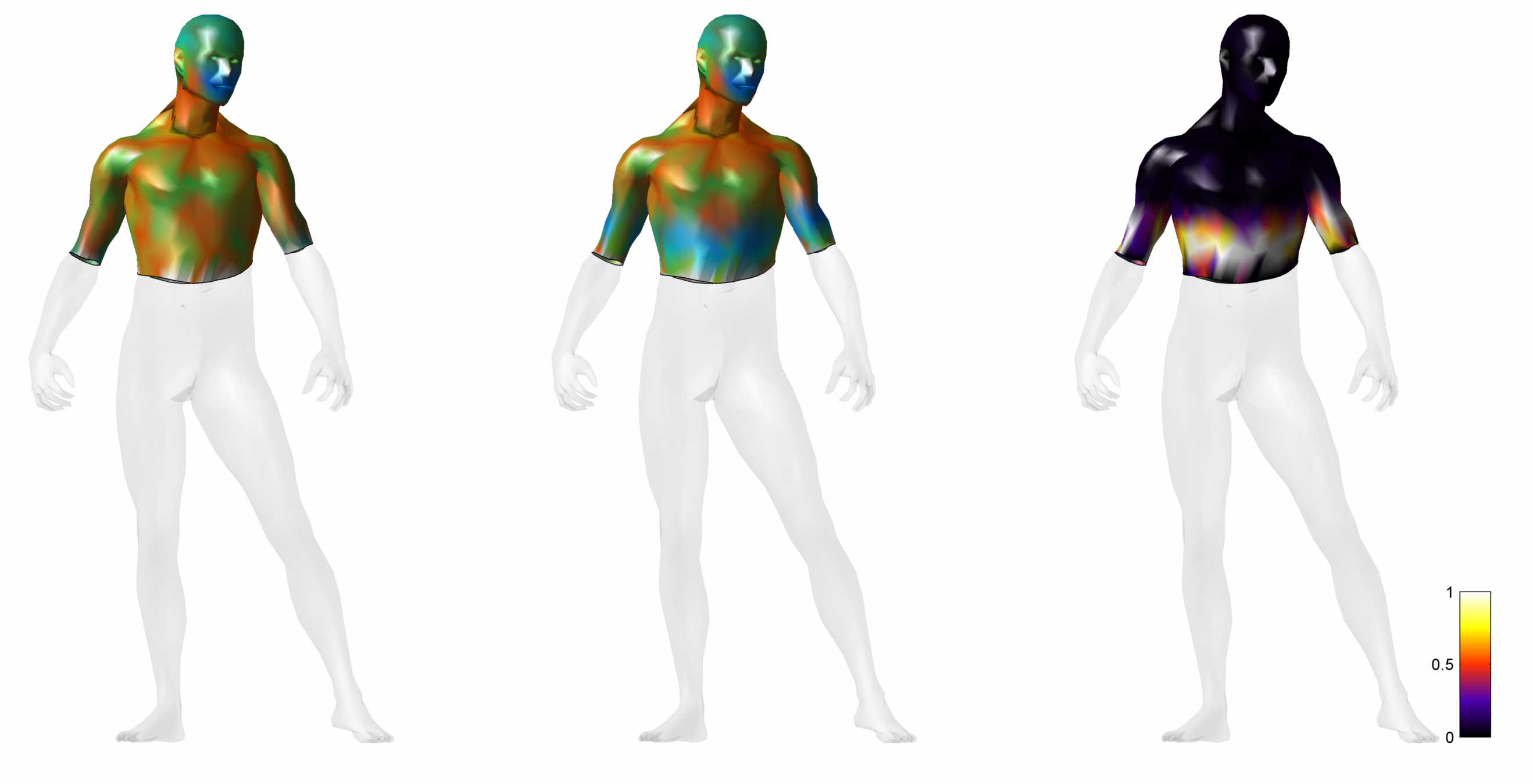}
    \end{center}
\end{small}
\caption{\label{fig:desc}\small{ An RGB visualization of the first three component of HKS descriptors computed on the full shape (left) and on a part of the shape (center). The $L_2$ difference between the two
fields is depicted in the rightmost figure. Note that the difference is maximal on the boundary decaying fast away from it; the error decay speed depends on the scale choice in the HKS descriptor. }}
\label{fig:edgeHKS}
\end{figure}

\begin{figure}[tbh]
	\centering
	\begin{minipage}[t]{\linewidth}
		\centering
			\includegraphics[width=0.8\linewidth]{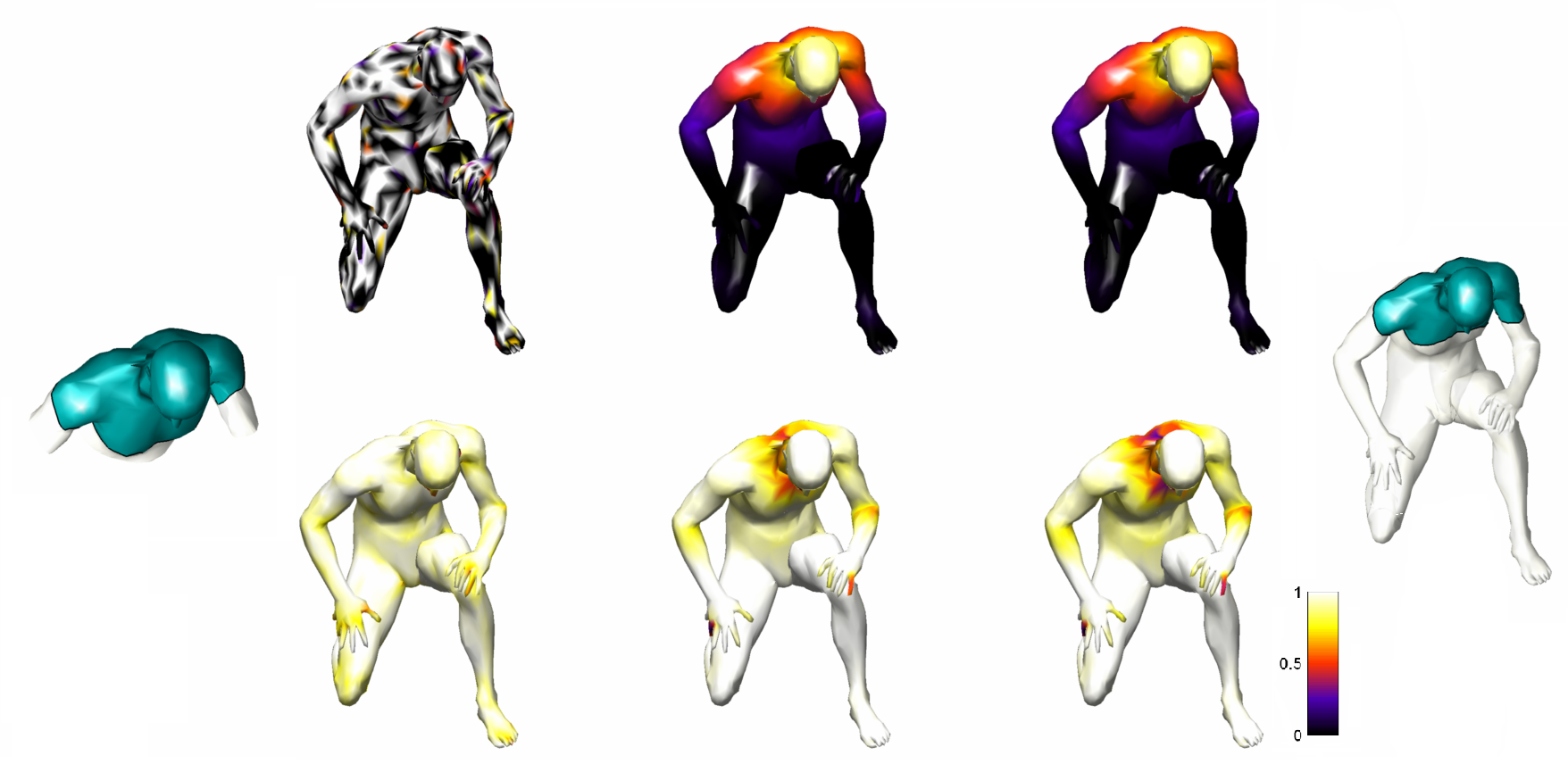}
			\caption{\label{fig:conv}\small{Convergence of the alternating minimization procedure. Depicted are the membership function $\uu$ (top row) and the phase field $\rho$ (bottom row) at the first three iterations.  }}
	\end{minipage}%
	\\
	\begin{minipage}[t]{\linewidth}
		\centering
			\includegraphics[width=\linewidth]{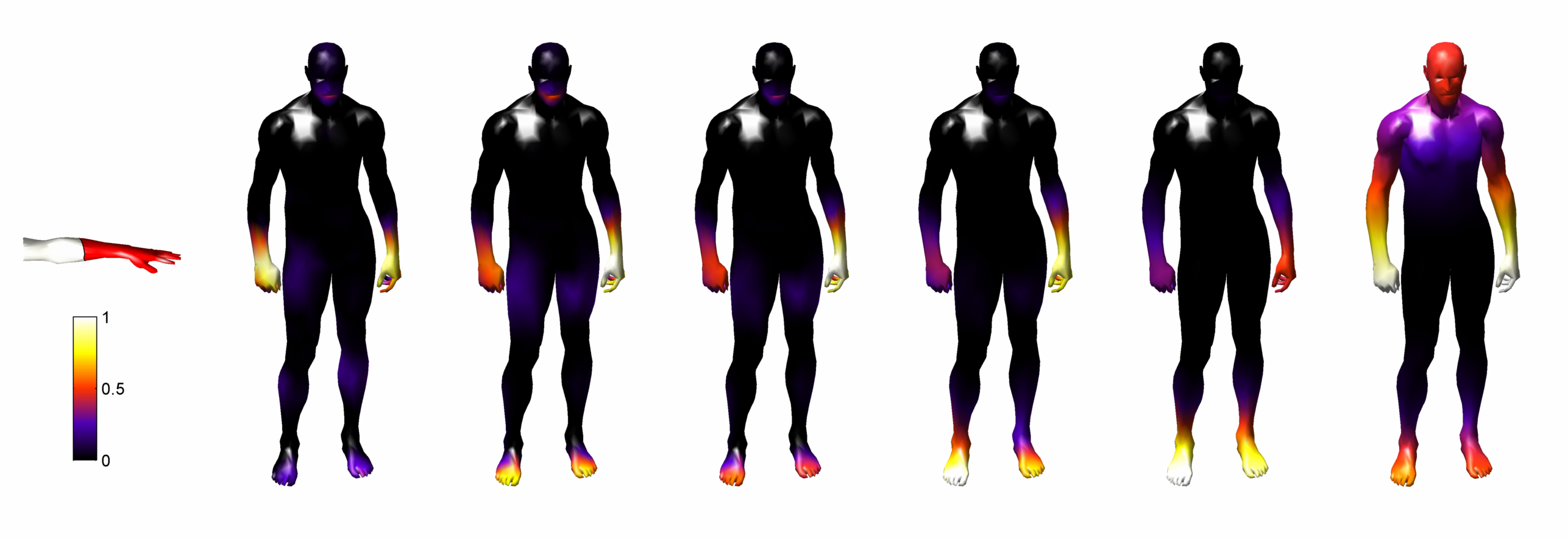}
\caption{\label{fig:param}\small{The influence of the parameter $\lambda_\mathrm{r}$, controlling the impact of regularization.
The leftmost figure depicts a query part; figures on the right are the membership function $u$ for different values of $\lambda_\mathrm{r}$. }}
\end{minipage}
\end{figure}

\begin{figure}[tbh]
	\centering
	\begin{minipage}[t]{\linewidth}
		\centering
	  \includegraphics[width=\linewidth]{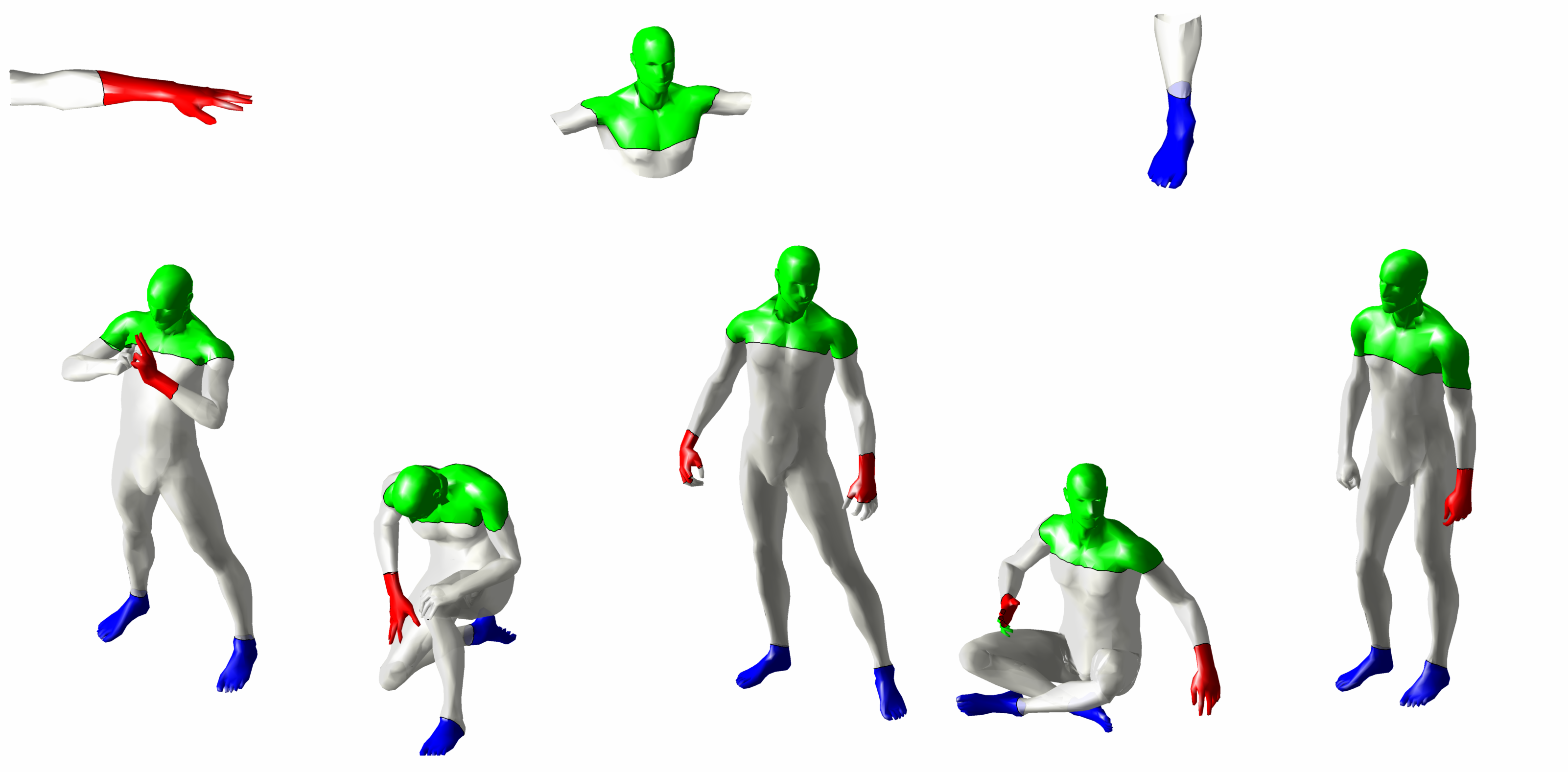}
	\caption{\label{fig:parts1}\small{Examples of matching of random parts of shapes (first row) to approximately isometric deformations of the shapes (second row). Color code indicates different parts. }}
	\end{minipage}%
 \\
	\begin{minipage}[t]{\linewidth}
		\centering
		\includegraphics[width=0.8\linewidth]{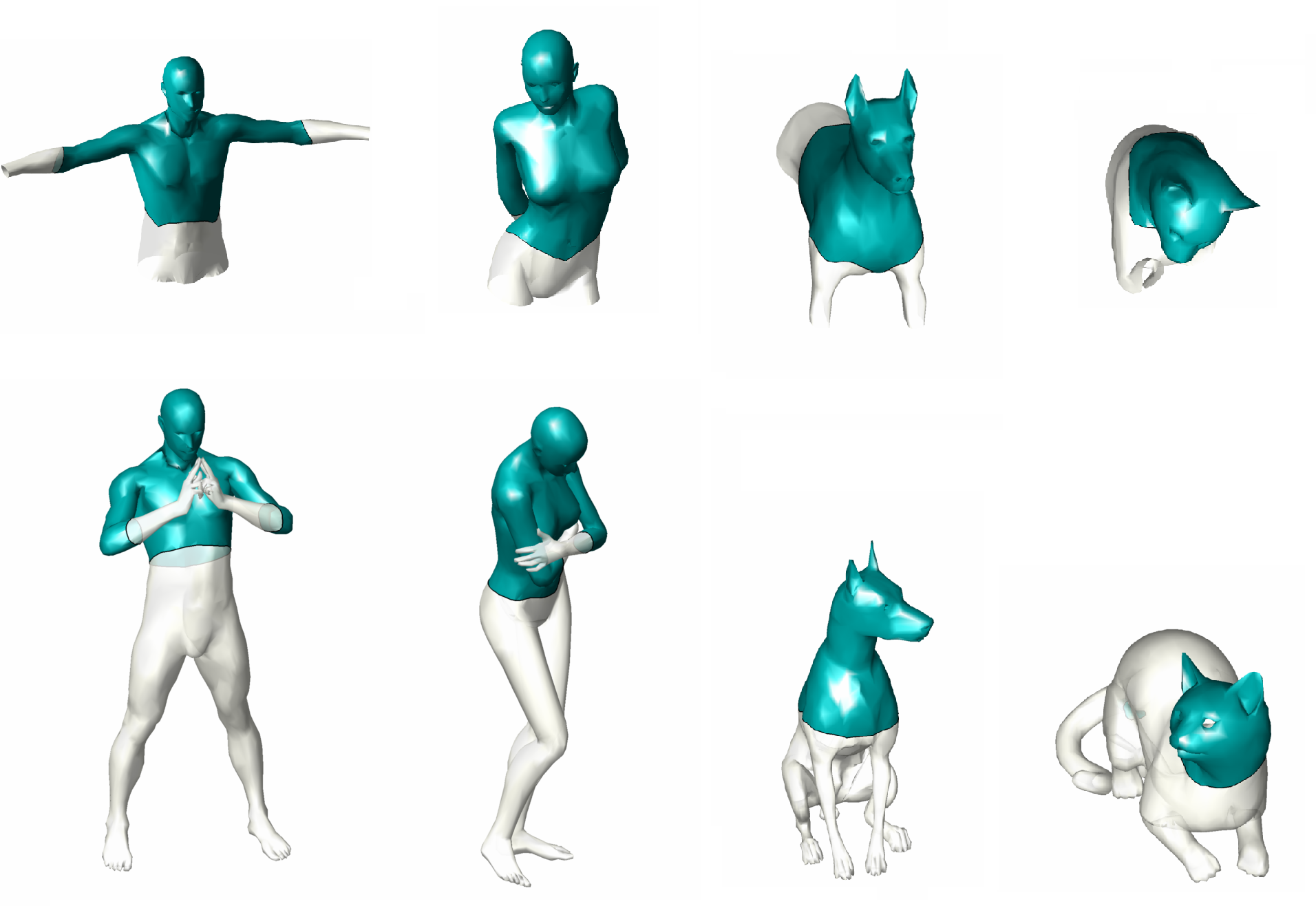}

		\caption{\label{fig:parts2}\small{Examples of matching of random parts of shapes (first row) to to approximately isometric deformations of the shapes (second row).}}

	\end{minipage}%
	\\
	\begin{minipage}[t]{\linewidth}
		\centering
		\includegraphics[width=0.75\linewidth]{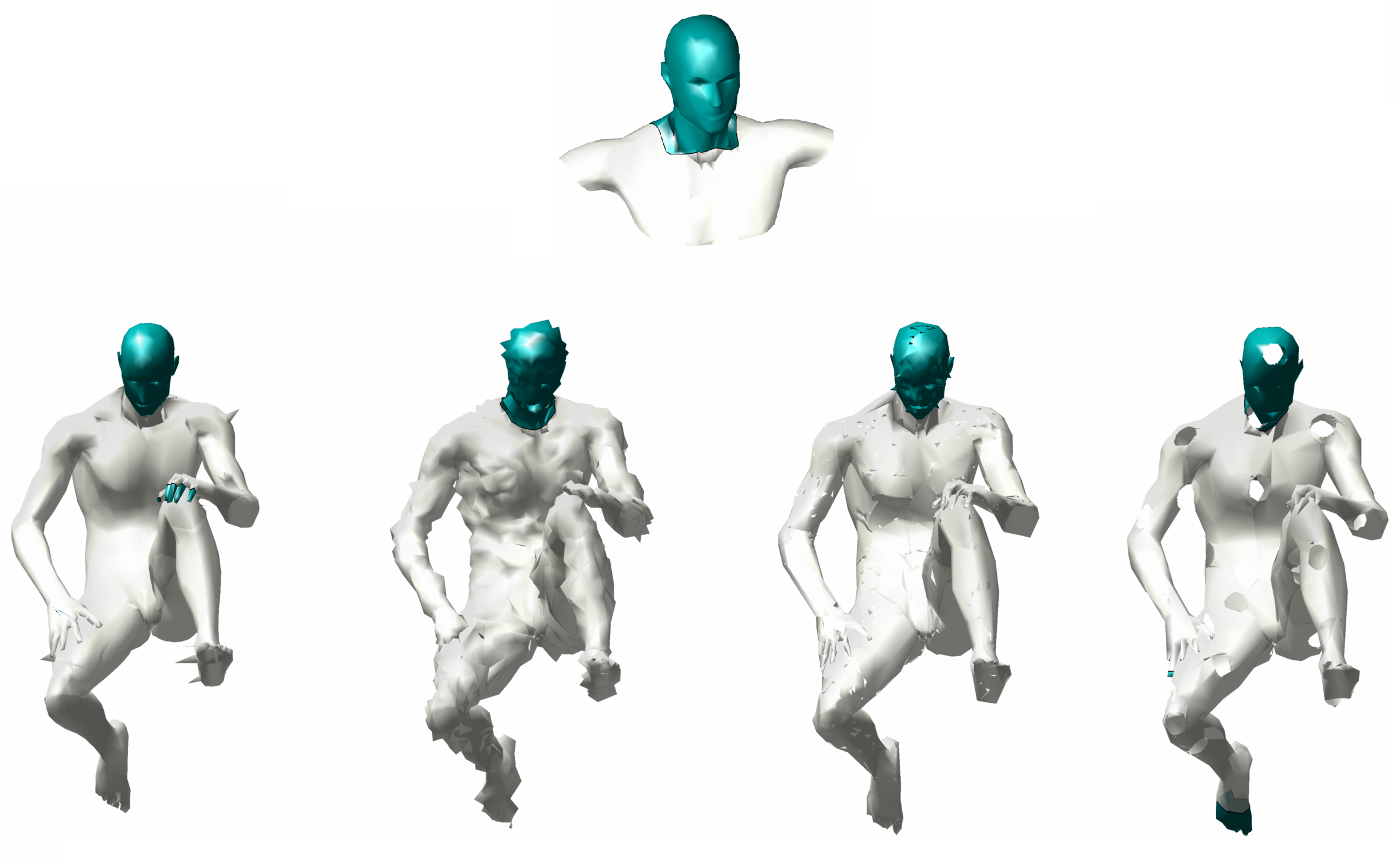}

\caption{\label{fig:parts3}\small{Results of matching a part of shape (first row) to shapes distorted by different transformations (second row). Shown left-to-right are: shot noise, noise, micro holes and holes.}}

	\end{minipage}%
\end{figure}

For each part, the normalized HKS descriptor was calculated at each vertex belonging to the part.
To avoid boundary effects (see Figure~\ref{fig:desc}),
descriptors close to the boundary were ignored when calculating $\overline{\qq}$ in (\ref{eq:orig-min-prob}).
The distance from the boundary was selected in accordance to the time scales of the HKS.
We used ten linearly spread samples in range $[65, 90]$ for the descriptors
and the according distance taken from edge was set to $15$.
Two to three iterations of the alternating minimization procedure were used, exhibiting fast
convergence (Figure~\ref{fig:conv}). After three iterations
the member function $\uu$ typically ceased changing significantly.
The phase map $\rho$ assumed the values close to $1$ in places of low gradient of the membership function $\uu$, and less than $1$ in high gradient areas
(Figure~\ref{fig:conv}). The importance of the regularization step is is evident observing the change in $\uu$ in Figure~\ref{fig:conv}.
Figure~\ref{fig:param} shows the influence of the parameter $\lambda_\mathrm{r}$,
controlling the impact of the regularization.
For too small values of $\lambda_\mathrm{r}$, two equally weighted
matches are obtained due to symmetry (left).
The phenomenon decreases with the increase of the influence of the regularization penalty.
However, increasing $\lambda_\mathrm{r}$ more causes incorrect matching (second and third columns from the right)
due to low data term influence.
Increasing it even more starts smoothing the result (rightmost column)
until eventually making the membership function uniform over the entire shape.
The resulting membership function $\uu$ was thresholded in such a way that the outcome area will be as close
as possible to the query area.
Figures~\ref{fig:parts1}--\ref{fig:parts3} show examples of matching results after thresholding.
Notice that in Figure~\ref{fig:parts1} the matching result sometimes contain the symmetric counter part of the result 
due to invariance of the HKS descriptor to intrinsic symmetry. 
(in this figure, the threshold was adjusted to the value of $0.35$ when the membership functions weights
are split between two symmetric parts as in Figure ~\ref{fig:param}).
The method is robust to shape deformations and geometric and topological noise as depicted in Figure~\ref{fig:parts3}.
Note that the figures show part-to-whole shape matching, but because of the low scale
HKS descriptors the same procedure works for matching to other parts as well.
Table~\ref{tab:overlap} summarizes quantitative evaluation that was performed on a
subset of the SHREC database, for which groundtruth correspondence and its bilaterally symmetric counterpart were available.
This subset included a male, a dog and a horse shape classes with different geometric, topological and noise deformations ($98$ shapes in total).
The query set was generated by selecting a part from a deformed shape ($1000$ queries in each deformation category)
 and matched to the null shape with parameters and thresholds as described above.
\begin{table}[tbh]
	\caption{\label{tab:overlap} \small{Part matching performance on transformed shapes from the SHREC benchmark.
At each query a random part (location and size) was selected from a deformed shape and matched to the null shape.
Overlap is reported compared to the groundtruth correspondence between the shapes (in parentheses taking into account the intrinsic bilateral
symmetry).}}
	\begin{center}
		\begin{tabular}{lcccc }
			\hline
			\hline
			Transformation  & \,\, &	Queries & \,\, &	Avg. overlap \\
			\hline\hline
			Isometry                       &  &   1000                   & &  75\%          (85\%)   \\
			Isometry + Shotnoise           &  &   1000                   & &  75\%          (85\%)  \\
			Isometry + Noise               &  &   1000                   & &  71\%          (82\%)  \\
			Isometry + Microholes          &  &   1000                   & &   68\%         (82\%) \\
			Isometry + Holes               &  &   1000                   & &  66\%          (76\%) \\
			\hline
			All                            &  &      5000                & &  71\%          (82\%)\\
			\hline \hline
		\end{tabular}
	\end{center}
\end{table}

\mypara{Complexity} The code was implemented in Matalb with some parts written in C with MEX interface.
The quadratic programming problem (\ref{eq:altmin-cont-step1}) in Step 1 was solved using QPC\footnote{ available online at \url{http://sigpromu.org/quadprog}} implementation of a dual active set method.
The experiments were run on 2.3GHz Intel Core2 Quad CPU, 2GB RAM in Win7 32bit environment. The running time (including re-meshing and descriptor calculation) per part was $40-50$ sec.

\section{Conclusions}


We presented a framework for finding partial similarity between shapes which does not rely on explicit correspondence. The method is based on regularized matching of region-wise local descriptors, and can be efficiently implemented. 
Experimental results show that our approach performs well in challenging matching scenarios, such as the presence of geometric and topological noise.
In the future work, we will extend the method to the setting of two partially-similar full shapes, in which two similar parts have to be found in each shape, and then consider  a multi-part matching (puzzle) scenario.

\bibliographystyle{plain}
\bibliography{ssvm11,bib_art}

\end{document}